\PassOptionsToPackage{unicode}{hyperref}
\PassOptionsToPackage{hyphens}{url}
\PassOptionsToPackage{dvipsnames,svgnames,x11names}{xcolor}
\documentclass[
]{article}

\usepackage{amsmath,amssymb}
\usepackage{iftex}
\ifPDFTeX
  \usepackage[T1]{fontenc}
  \usepackage[utf8]{inputenc}
  \usepackage{textcomp} 
\else 
  \usepackage{unicode-math}
  \defaultfontfeatures{Scale=MatchLowercase}
  \defaultfontfeatures[\rmfamily]{Ligatures=TeX,Scale=1}
\fi
\usepackage{lmodern}
\ifPDFTeX\else  
\fi
\IfFileExists{upquote.sty}{\usepackage{upquote}}{}
\IfFileExists{microtype.sty}{
  \usepackage[]{microtype}
  \UseMicrotypeSet[protrusion]{basicmath} 
}{}
\makeatletter
\@ifundefined{KOMAClassName}{
  \IfFileExists{parskip.sty}{%
    \usepackage{parskip}
  }{
    \setlength{\parindent}{0pt}
    \setlength{\parskip}{6pt plus 2pt minus 1pt}}
}{
  \KOMAoptions{parskip=half}}
\makeatother
\usepackage{xcolor}
\makeatletter
\ifx\paragraph\undefined\else
  \let\oldparagraph\paragraph
  \renewcommand{\paragraph}{
    \@ifstar
      \xxxParagraphStar
      \xxxParagraphNoStar
  }
  \newcommand{\xxxParagraphStar}[1]{\oldparagraph*{#1}\mbox{}}
  \newcommand{\xxxParagraphNoStar}[1]{\oldparagraph{#1}\mbox{}}
\fi
\ifx\subparagraph\undefined\else
  \let\oldsubparagraph\subparagraph
  \renewcommand{\subparagraph}{
    \@ifstar
      \xxxSubParagraphStar
      \xxxSubParagraphNoStar
  }
  \newcommand{\xxxSubParagraphStar}[1]{\oldsubparagraph*{#1}\mbox{}}
  \newcommand{\xxxSubParagraphNoStar}[1]{\oldsubparagraph{#1}\mbox{}}
\fi
\makeatother

\providecommand{\tightlist}{%
  \setlength{\itemsep}{0pt}\setlength{\parskip}{0pt}}\usepackage{longtable,booktabs,array}
\usepackage{calc} 
\usepackage{etoolbox}
\makeatletter
\patchcmd\longtable{\par}{\if@noskipsec\mbox{}\fi\par}{}{}
\makeatother
\IfFileExists{footnotehyper.sty}{\usepackage{footnotehyper}}{\usepackage{footnote}}
\makesavenoteenv{longtable}
\usepackage{graphicx}
\makeatletter
\newsavebox\pandoc@box
\newcommand*\pandocbounded[1]{
  \sbox\pandoc@box{#1}%
  \Gscale@div\@tempa{\textheight}{\dimexpr\ht\pandoc@box+\dp\pandoc@box\relax}%
  \Gscale@div\@tempb{\linewidth}{\wd\pandoc@box}%
  \ifdim\@tempb\p@<\@tempa\p@\let\@tempa\@tempb\fi
  \ifdim\@tempa\p@<\p@\scalebox{\@tempa}{\usebox\pandoc@box}%
  \else\usebox{\pandoc@box}%
  \fi%
}
\def\fps@figure{htbp}
\makeatother
\NewDocumentCommand\citeproctext{}{}

\makeatletter
 \let\@cite@ofmt\@firstofone
 \def\@biblabel#1{}
 \def\@cite#1#2{{#1\if@tempswa , #2\fi}}
\makeatother
\newlength{\cslhangindent}
\newlength{\csllabelwidth}
\newenvironment{CSLReferences}[2] 
 {\begin{list}{}{%
  \setlength{\itemindent}{0pt}
  \setlength{\leftmargin}{0pt}
  \setlength{\parsep}{0pt}
  \ifodd #1
   \setlength{\leftmargin}{\cslhangindent}
   \setlength{\itemindent}{-1\cslhangindent}
  \fi
  \setlength{\itemsep}{#2\baselineskip}}}
 {\end{list}}
\usepackage{calc}

\usepackage{arxiv}
\usepackage{orcidlink}
\usepackage{amsmath}
\usepackage[T1]{fontenc}
\makeatletter
\@ifpackageloaded{caption}{}{\usepackage{caption}}
\AtBeginDocument{%
\ifdefined\contentsname
  \renewcommand*\contentsname{Table of contents}
\else
  \newcommand\contentsname{Table of contents}
\fi
\ifdefined\listfigurename
  \renewcommand*\listfigurename{List of Figures}
\else
  \newcommand\listfigurename{List of Figures}
\fi
\ifdefined\listtablename
  \renewcommand*\listtablename{List of Tables}
\else
  \newcommand\listtablename{List of Tables}
\fi
\ifdefined\figurename
  \renewcommand*\figurename{Figure}
\else
  \newcommand\figurename{Figure}
\fi
\ifdefined\tablename
  \renewcommand*\tablename{Table}
\else
  \newcommand\tablename{Table}
\fi
}
\@ifpackageloaded{float}{}{\usepackage{float}}
\floatstyle{ruled}
\@ifundefined{c@chapter}{\newfloat{codelisting}{h}{lop}}{\newfloat{codelisting}{h}{lop}[chapter]}
\floatname{codelisting}{Listing}

\makeatother
\makeatletter
\@ifpackageloaded{caption}{}{\usepackage{caption}}
\@ifpackageloaded{subcaption}{}{\usepackage{subcaption}}
\makeatother

\usepackage{bookmark}

\IfFileExists{xurl.sty}{\usepackage{xurl}}{} 
\hypersetup{
  pdftitle={PyPotteryLens: An Open-Source Deep Learning Framework for Automated Digitisation of Archaeological Pottery Documentation},
  pdfauthor={Lorenzo Cardarelli},
  pdfkeywords={Deep Learning, Pottery, Legacy Data, Archaeology},
  colorlinks=true,
  linkcolor={blue},
  filecolor={Maroon},
  citecolor={Blue},
  urlcolor={Blue},
  pdfcreator={LaTeX via pandoc}}

\newcommand{\runninghead}{A Preprint }
\title{\emph{PyPotteryLens}: An Open-Source Deep Learning Framework for
Automated Digitisation of Archaeological Pottery Documentation}
\author{\textbf{Lorenzo
Cardarelli}~\orcidlink{0000-0002-2436-9967}\\Department of
Antiquities\\Sapienza University of Rome\\Rome,
RM,\ 135\\\href{mailto:lorenzo.cardarelli@uniroma1.it}{lorenzo.cardarelli@uniroma1.it}}
\date{}
\begin{document}
\maketitle
\begin{abstract}
Archaeological pottery documentation and study represents a crucial
but time-consuming aspect of archaeology. While recent years have seen
advances in digital documentation methods, vast amounts of legacy data
remain locked in traditional publications. This paper introduces
\emph{PyPotteryLens}, an open-source framework that leverages deep
learning to automate the digitisation and processing of archaeological
pottery drawings from published sources. The system combines
state-of-the-art computer vision models (YOLO for instance segmentation
and EfficientNetV2 for classification) with an intuitive user interface,
making advanced digital methods accessible to archaeologists regardless
of technical expertise. The framework achieves over 97\% precision and
recall in pottery detection and classification tasks, while reducing
processing time by up to 5× to 20× compared to manual methods. Testing
across diverse archaeological contexts demonstrates robust
generalisation capabilities. Also, the system's modular architecture
facilitates extension to other archaeological materials, while its
standardised output format ensures long-term preservation and
reusability of digitised data as well as solid basis for training
machine learning algorithms. The software, documentation, and examples
are available on GitHub
(https://github.com/lrncrd/PyPottery/tree/PyPotteryLens).
\end{abstract}
{\bfseries \emph Keywords}
\def\sep{\textbullet\ }
Deep Learning \sep Pottery \sep Legacy Data \sep 
Archaeology

\section{Introduction}\label{introduction}

Working with archaeological pottery is a time-consuming and
resource-intensive process. Once unearthed, after a process of cleaning
and cataloguing, archaeological ceramics are traditionally documented
through technical drawings, which are then illustrated as plates or
tables in publications. While new methods of digital documentation have
been developed in recent years, allowing speed-up and standardisation of
the drawing process (Demján, Pavúk, and Roosevelt 2023), there are
hundreds of thousands of data already available in paper publications
(or PDFs) that can be considered as \emph{legacy data} (Allison 2008;
Snow 2010). As archaeological ceramics are a key element in dating and
interpreting archaeological contexts (Shepard 1985; Orton and Hughes
2013; Sinopoli 1991), it is important that this data is made digitally
accessible and re-usable (Marwick et al. 2017). Traditionally, the
retrieval of data from publications has involved a lengthy process of
manual digitisation, starting with the extraction of the drawings using
photo-editing software, which is then associated with a range of tabular
information, usually managed in a spreadsheet or database. Clearly, this
is a tedious, time-consuming and error-prone process, and it can become
a particularly challenging task when dealing with very large datasets.
Nevertheless, archaeological drawings represent a source of knowledge
that is essentially (or highly) standardised as they contain information
on the shape, decoration, and dimensions of artefacts and provide a
highly analytical value that can be used in traditional or typological
approaches (Clarke 1968; Adams and Adams 1991; Read 2009) and modern
computational methods, including the application of machine learning
(ML) (Navarro et al. 2021, 2022; Parisotto et al. 2022; L. Cardarelli
2022). Recently, the use of deep learning (DL) techniques has become
increasingly popular in archaeology (Bickler 2021; Cacciari and
Pocobelli 2022), following trends in other fields, including
non-scientific ones, such as arts or industries (Le et al. 2020; Ramesh
et al. 2021; Maerten and Soydaner 2024). However, many applications have
focused on specific research areas like site detection (Sakai et al.
2024; Caspari and Crespo 2019; Buławka, Orengo, and Berganzo-Besga
2024), artifact classification (Emmitt et al. 2022; Gualandi, Gattiglia,
and Anichini 2021; Anichini et al. 2021) or heritage preservation (Cui
et al. 2024; D'Orazio et al. 2024), while the retrieval and processing
of \emph{legacy data} has remained relatively unexplored. This gap is
particularly significant given that the efficient and rapid retrieval of
archaeological data from printed publications or PDFs is essential for
the advancement of archaeological research, as we can use this
ready-to-use data for both traditional analysis and the development of
even more complex ML techniques that require even larger training
datasets.

This paper presents \emph{PyPotteryLens}, a specialised framework
designed to transform legacy pottery documentation into standardised
digital records while preserving scientific or analytical strength. The
approach embraces DL techniques and makes three principal key
contributions to pottery studies and digital heritage preservation:

\begin{itemize}
\tightlist
\item
  A robust methodology for automatically detecting and segmenting
  pottery drawings from publications, creating high-quality digital
  records suitable for long-term preservation.
\item
  A classification system that standardises the orientation and
  preservation of pottery records, ensuring consistency across digital
  archives and prepare the data for the subsequent analysis.
\item
  An interactive visualisation framework that enables users' examination
  and validation providing accuracy in complex cases or errors.
\end{itemize}

The program requires no specific programming skills as it combines
powerful DL techniques with an intuitive, easy-to-use, and scalable user
interface. Totally open-source, the software is accessible, improvable
and expandable by the archaeological community. In fact, although the
system is designed for the recording of archaeological ceramics, it can
be easily adapted for the analysis and documentation of other types of
archaeological objects such as lithics, metals and so on.

\begin{figure}[!b]

\centering{

\includegraphics[width=0.8\linewidth,height=\textheight,keepaspectratio]{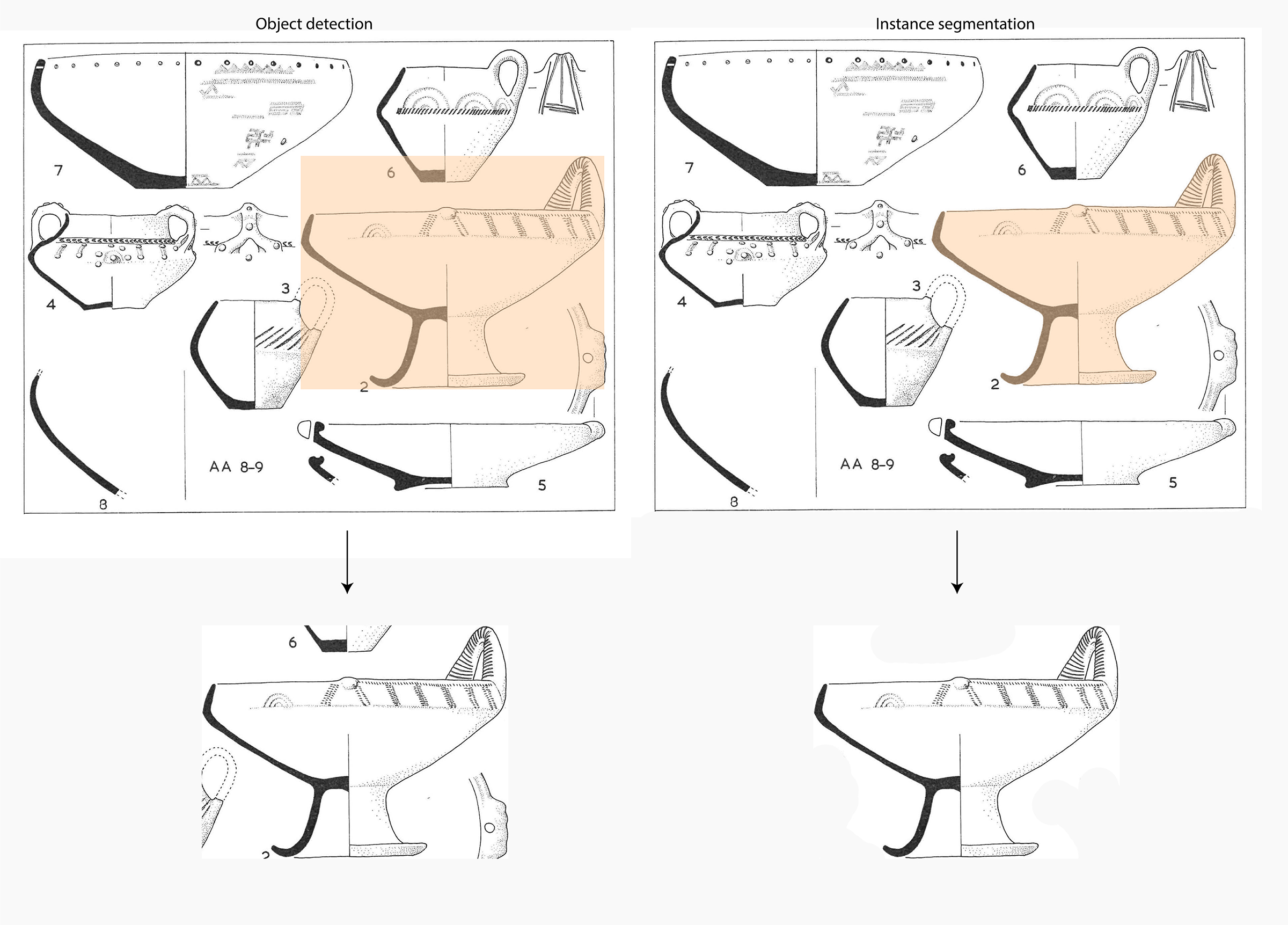}

}

\caption{\label{fig-seg}Object detection (left) \emph{versus} instance
segmentation (right). The example illustrates the principal benefits of
the segmentation approach, which enables the removal of each element
that falls outside the target area. Example by Moretti et al. (1963).}

\end{figure}

Recent approaches to documenting archaeological material include work by
Klein et al. (2024), who propose a method for analysing PDFs using
computer vision techniques. Their method focuses on object detection
across 14 classes of archaeological objects, including ceramics, using
the ResNet-152 neural network (He et al. 2015). In this paper, a
different approach is suggested: first, it propose a more vertical
approach as it specialises in ceramics, allowing for more precise and
targeted analysis; second, it employs the YOLO architecture, which
offers significant advantages in processing speed and efficiency
compared to ResNet-152. Most importantly, the proposed method implements
instance segmentation rather than simple object detection, enabling
precise contour tracing of pottery drawings (Figure~\ref{fig-seg}). This
segmentation-based approach provides several advantages: it allows for
more accurate measurements, facilitates detailed shape analysis, and
produces cleaner results suitable for both digital archives and further
computational analysis. Additionally, the segmentation masks generated
can be used to create new training data, enabling continuous improvement
of the model's performance and adaptability to different pottery styles
and publication formats through a self-training approach (Amini et al.
2024).

Finally, the proposed methodology includes extensive documentation and a
series of practical examples to facilitate the use of the software by
archaeologists. This approach is designed to ensure that the software is
accessible to all users, regardless of their technical background, and
to facilitate integration with traditional archaeological working
methods. The software, documentation and examples are available on
GitHub (\url{https://github.com/lrncrd/PyPottery/tree/PyPotteryLens}) to
ensure maximum transparency and accessibility for the archaeological
community and to encourage collaboration and further development.

\section{Research aims}\label{research-aims}

This paper provides a comprehensive analysis of the \emph{PyPotteryLens}
pipeline, with three key research objectives:

\begin{enumerate}
\def\labelenumi{\arabic{enumi}.}
\item
  To introduce \emph{PyPotteryLens} as a solution for automated pottery
  recording in archaeology, specifically: (1) Demonstrating its capacity
  to process and digitise legacy archaeological publications (2)
  Evaluating its effectiveness in reducing processing time compared to
  manual methods.
\item
  To present a detailed technical examination of the system
  architecture, focusing on (1) The implementation and performance of
  YOLO-based instance segmentation for pottery detection (2) The
  development and validation of a custom EfficientNetV2-based
  classification model for drawing orientation and fragmentation
  identification (3) Testing Gradio user interface as integration of DL
  models within a user-friendly interface.
\item
  To evaluate the system's effectiveness through (1) Comprehensive
  testing on a diverse corpus of archaeological publications spanning
  different chronological periods and regions (2) Quantitative
  assessment of accuracy, precision, and recall rates for both detection
  and classification tasks (3) Analysis of the system's ability to
  handle variations in publication formats and drawing styles (4)
  Demonstration of the system's potential for generating standardised
  training data for future deep learning applications in archaeology.
\end{enumerate}

\section{Methods}\label{methods}

\subsection{General system
architecture}\label{general-system-architecture}

\emph{PyPotteryLens} is structured as a comprehensive framework for
analysing archaeological pottery contained within PDF documents or
images. At its core, the system integrates six main modules that work in
concert: \emph{Document Processing}, \emph{Model Processing}, \emph{Mask
Extraction}, \emph{Tabular Data Management}, \emph{Post-processing} and
\emph{Self-Annotation}.

\begin{figure}

\centering{

\includegraphics[width=0.6\linewidth,height=\textheight,keepaspectratio]{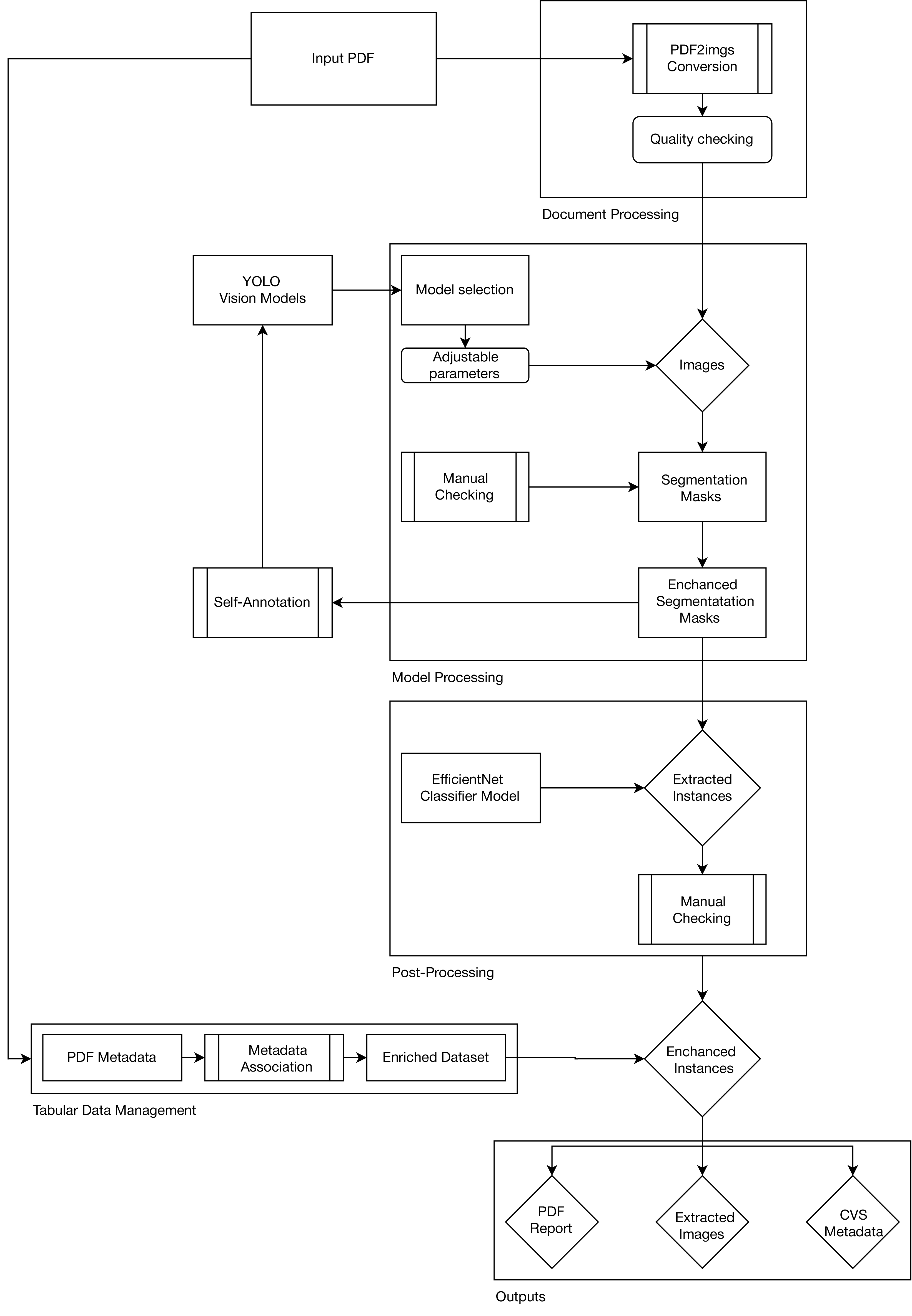}

}

\caption{\label{fig-diagram}A schematic diagram shows the workflow of
the proposed system. Made with
\href{https://app.diagrams.net/}{draw.io}.}

\end{figure}

The system's workflow progresses through several interconnected phases
(Figure~\ref{fig-diagram}), beginning with the \emph{Document
Processing} stage where PDF documents are fed into the system. These
inputs undergo initial processing and normalisation, with an integrated
visualiser allowing users to directly assess image quality within the
application. Moving into the \emph{Model Processing} stage, the system
applies YOLO model inference using a set of user-adjustable parameters,
providing flexibility to accommodate varying document qualities and
pottery styles (e.g., post-processing on segmentation masks or
confidence threshold for the model). The resulting segmentation masks
can be visually modified or supplemented through manual control,
ensuring high accuracy in complex cases or errors. If the model fails to
detect instances, new masks can be created manually. Moving to
\emph{Mask Extraction} step, each identified pottery instance is then
extracted and established as an individual \emph{.png} file, creating
\emph{discrete units} for subsequent analysis and processing. The
\emph{Tabular Data Management} enables the association of detailed
metadata with each extracted instance, creating a dataset that combines
image and contextual information through the creation of a spreadsheet
(\emph{.csv}) containing all the inserted information. In the
\emph{Post-processing} stage, \emph{PyPotteryLens} employs a custom
EfficientNetV2 (Tan and Le 2021) model to classify the extracted
instances according to predefined categories. To ensure high
performance, these classifications can be visualised and manually
adjusted if needed. After this step, the recording process is completed,
and the results can be indexed and exported as images and in optimised
PDF format using a two-dimensional bin packing algorithm (Lodi,
Martello, and Vigo 2002).

Finally, the \emph{Self-Annotation} module allows users transform the
extracted mask into a new fuel for training the computer vision model,
creating a virtuous cycle of data generation and model improvement based
on your specific needs.

\subsection{Deep Learning
Implementation}\label{deep-learning-implementation}

The application uses two deep learning models for image analysis: YOLO
for image recognition and segmentation, and a custom EfficientNetV2 for
classification.

\subsubsection{YOLO Model Architecture and Training
Dataset}\label{yolo-model-architecture-and-training-dataset}

\begin{figure}

\centering{

\includegraphics[width=0.5\linewidth,height=\textheight,keepaspectratio]{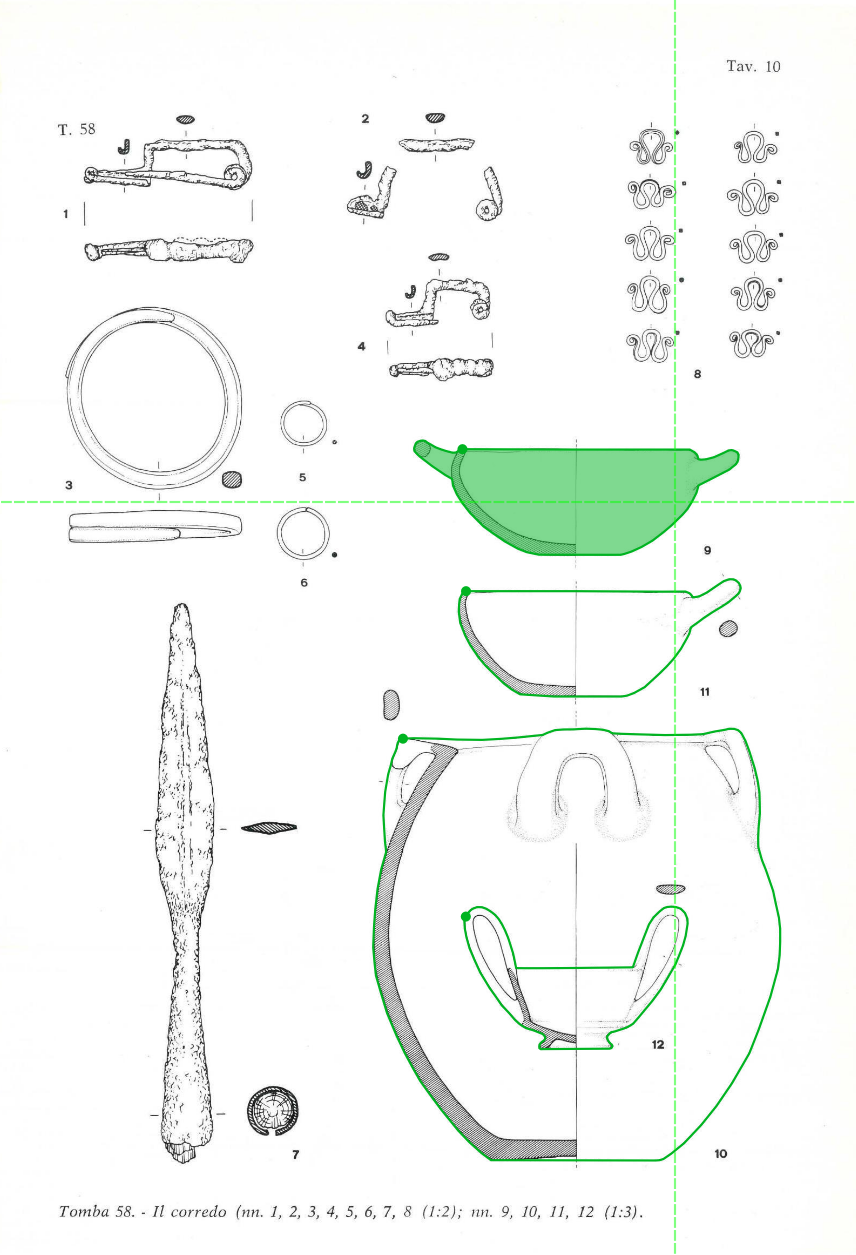}

}

\caption{\label{fig-labelme}An annotated page by Parise Badoni and
Ruggeri Giove (1980) using the software \emph{LabelMe}.}

\end{figure}

\begin{figure}

\centering{

\pandocbounded{\includegraphics[keepaspectratio]{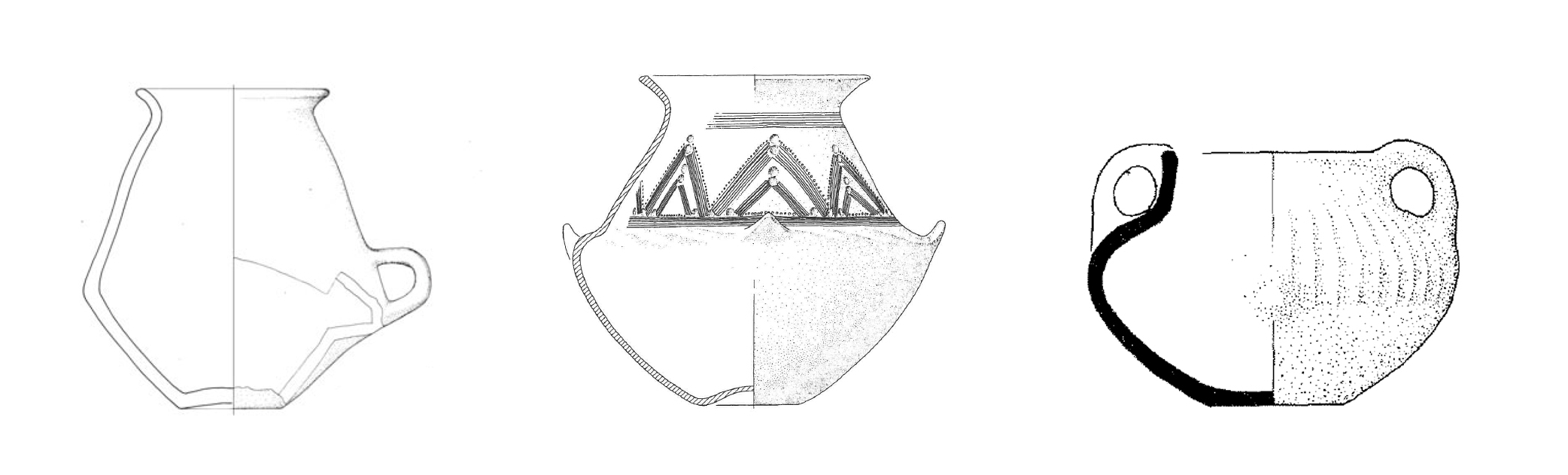}}

}

\caption{\label{fig-pottery_style}Some examples of the styles used in
the representation of archaeological ceramics. From left to right:
Bianco Peroni, Peroni, and Vanzetti (2010) pag. 105, tab. 62, AI; L.
Salzani and Colonna (2010) pag. 326, tav. 11, B1; Moretti et al. (1963)
pag. 108; tav. 36, c.}

\end{figure}

YOLO (Redmon et al. 2016; Jocher, Qiu, and Chaurasia 2023) is a popular
open-source computer vision framework that includes a set of pre-trained
models and architecture that can be used for various image analysis
tasks, including object detection, instance segmentation, and pose
detection. In this analysis, two models, YOLOv8 and YOLOv11, are
compared to evaluate their performance. In both cases, the \emph{Xlarge}
version is used, which sacrifices execution speed for greater accuracy.
The models are fine-tuned on a dataset of 4097 manually annotated
archaeological pottery instances from 13 different publications from
settlements and necropolises of the 2\textsuperscript{nd} and
1\textsuperscript{st} millennium BCE in continental Italy, covering a
wide range of ceramic styles or types (Figure~\ref{fig-labelme}). In
addition to the nature of the pottery itself, which can vary greatly in
terms of shape, size and decoration, the publication of these data is
faced with several stylistic and technical choices that pose a challenge
to automatic analysis: for example, the profiles of the vessels can be
sampled in different ways, as can the graphic rendering of the
prospectus, especially shadows and decorations
(Figure~\ref{fig-pottery_style}). It was therefore decided to use a
heterogeneous dataset to ensure maximum generalisation of the model.

As customary, 80\% of the images were used for training and the
remaining 20\% for validation. For both YOLOv8 and YOLOv11, an
automatically defined batch number is used by the model to optimise the
performance. In addition, although a maximum number of epochs is set at
200, an early stopping parameter is set to avoid overfitting the model.
The training configuration files are accessible on the GitHub
repository. Due to copyright restrictions, the training dataset is not
publicly available.

The table resume the dataset used for training the model
(Table~\ref{tbl-dataset}).

\begin{longtable}[]{@{}
  >{\raggedright\arraybackslash}p{(\linewidth - 6\tabcolsep) * \real{0.1167}}
  >{\raggedright\arraybackslash}p{(\linewidth - 6\tabcolsep) * \real{0.5500}}
  >{\raggedright\arraybackslash}p{(\linewidth - 6\tabcolsep) * \real{0.1833}}
  >{\raggedright\arraybackslash}p{(\linewidth - 6\tabcolsep) * \real{0.1500}}@{}}
\caption{The table shows the bibliographic references for the training
dataset, as well as the number of annotations and the annotator (LC:
Lorenzo Cardarelli; LP: Lucrezia Petrucci; AL: Annalisa Lapadula; DC:
Dacia Cardarelli)}\label{tbl-dataset}\tabularnewline
\toprule\noalign{}
\begin{minipage}[b]{\linewidth}\raggedright
ID
\end{minipage} & \begin{minipage}[b]{\linewidth}\raggedright
References
\end{minipage} & \begin{minipage}[b]{\linewidth}\raggedright
Annotations
\end{minipage} & \begin{minipage}[b]{\linewidth}\raggedright
Annotator
\end{minipage} \\
\midrule\noalign{}
\endfirsthead
\toprule\noalign{}
\begin{minipage}[b]{\linewidth}\raggedright
ID
\end{minipage} & \begin{minipage}[b]{\linewidth}\raggedright
References
\end{minipage} & \begin{minipage}[b]{\linewidth}\raggedright
Annotations
\end{minipage} & \begin{minipage}[b]{\linewidth}\raggedright
Annotator
\end{minipage} \\
\midrule\noalign{}
\endhead
\bottomrule\noalign{}
\endlastfoot
ALF & Parise Badoni and Ruggeri Giove (1980) & 136 & LC \\
CMP\_I & Chiaramonte Treré and D'Ercole (2003) & 276 & LC; LP \\
CMP\_II & Chiaramonte Treré et al. (2010) & 362 & LP \\
FIAV & Perini (1994) & 636 & LP \\
FOSS\_I & Cosentino, D'Ercole, and Mieli (2001) & 104 & AL \\
FOSS\_II & Cosentino, D'Ercole, and Mieli (2004) & 210 & AL \\
LRS & Buranelli (1983) & 170 & DC \\
NRD & L. Salzani and Colonna (2010) & 237 & LC \\
MRN & Giaretti and Rubat Borel (2006) & 109 & LC \\
PDG & Bianco Peroni, Peroni, and Vanzetti (2010) & 717 & LC \\
QF\_1963 & Moretti et al. (1963) & 306 & LC \\
QF\_1965 & Ward-Perkins et al. (1965) & 230 & LC \\
TRRGLL & Pacciarelli (1999) & 604 & LC \\
Total & & 4097 & \\
\end{longtable}

\subsubsection{Classifications Model and
Training}\label{classifications-model-and-training}

The second phase of the DL pipeline incorporates a classification model
designed to standardise the presentation and interpretation of ceramic
instances. This classification serves two primary purposes: (1)
determining whether the instance represents a complete vessel (ENT) or a
fragment (FRAG), and (2) standardising the orientation of each instance
to ensure consistent presentation with the vessel's mouth oriented
upward (TOP/BOTTOM) and its section positioned to the left (LEFT/RIGHT).

The model's training dataset includes different publications for a total
of 4563 instances\footnote{The publications used include Bietti Sestieri
  (1992), Weidig and Egg (2021), A. Cardarelli and Malnati (2003) in
  addition to the training dataset used for the YOLO model
  (Table~\ref{tbl-dataset}). Dataset is handled to ensure consistency
  between fragmentary and complete vessels.}, divided into the following
categories:

\begin{itemize}
\tightlist
\item
  2,428 fragmentary vessels
\item
  2,135 complete vessels
\end{itemize}

These base images are manually augmented to account for all possible
orientation combinations (e.g., FRAG-TOP-RIGHT, FRAG-TOP-LEFT),
resulting in 8 distinct configurations per instance and a comprehensive
training set of 18,252 examples. In this training, 80\% of the images
are used for training and the remaining 20\% for validation.

To effectively handle this multi-faceted classification task in a single
network, a multi head architecture based on EfficientNetV2 is
implemented. The multi head architecture consists of a shared backbone
(EfficientNetV2) that extracts general features from the input images
and three specialised classification heads, each dedicated to a specific
categorisation task: vessel completeness (Type: ENT/FRAG), vertical
orientation (Position: TOP/BOTTOM), and horizontal orientation
(Rotation: LEFT/RIGHT). The model uses a combined loss function that
aggregates the three individual losses to ensure balanced learning
across all three tasks. Each loss is defined as the binary cross-entropy
(Mao, Mohri, and Zhong 2023) between the prediction and the target from
each classification head.

Further details on the model architecture and training configuration are
available in the GitHub repository.

\subsubsection{\texorpdfstring{\emph{Self-Annotation}
Module}{Self-Annotation Module}}\label{sec-self-annotation}

\begin{figure}

\centering{

\pandocbounded{\includegraphics[keepaspectratio]{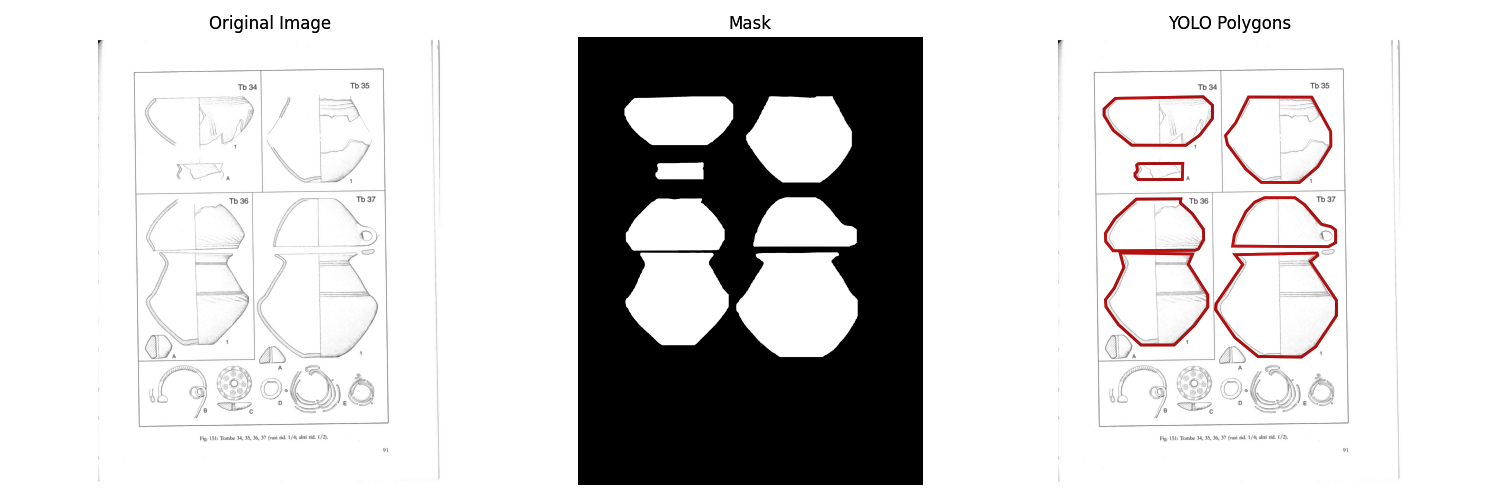}}

}

\caption{\label{fig-self-annots}A plot showing how new training data is
created by the \emph{Self-Annotation} module.}

\end{figure}

The \emph{Self-Annotation} module is designed to facilitate the
continuous improvement of the YOLO model through the generation of new
training data. Once the model is applied and the manual corrections are
made, the user can export the masks in a ready-to-use format for YOLO
training (Figure~\ref{fig-self-annots}). A script facilitates the
fine-tuning of the model with the new data. Due to the complexity of the
operation, this module is intended for users with a deeper understanding
of DL concepts, no user interface is provided.

You can find the code and the documentation on the GitHub repository.

\subsection{The User Interface}\label{the-user-interface}

\begin{figure}

  \centering{
  
  \pandocbounded{\includegraphics[width=0.6\linewidth,height=\textheight,keepaspectratio]{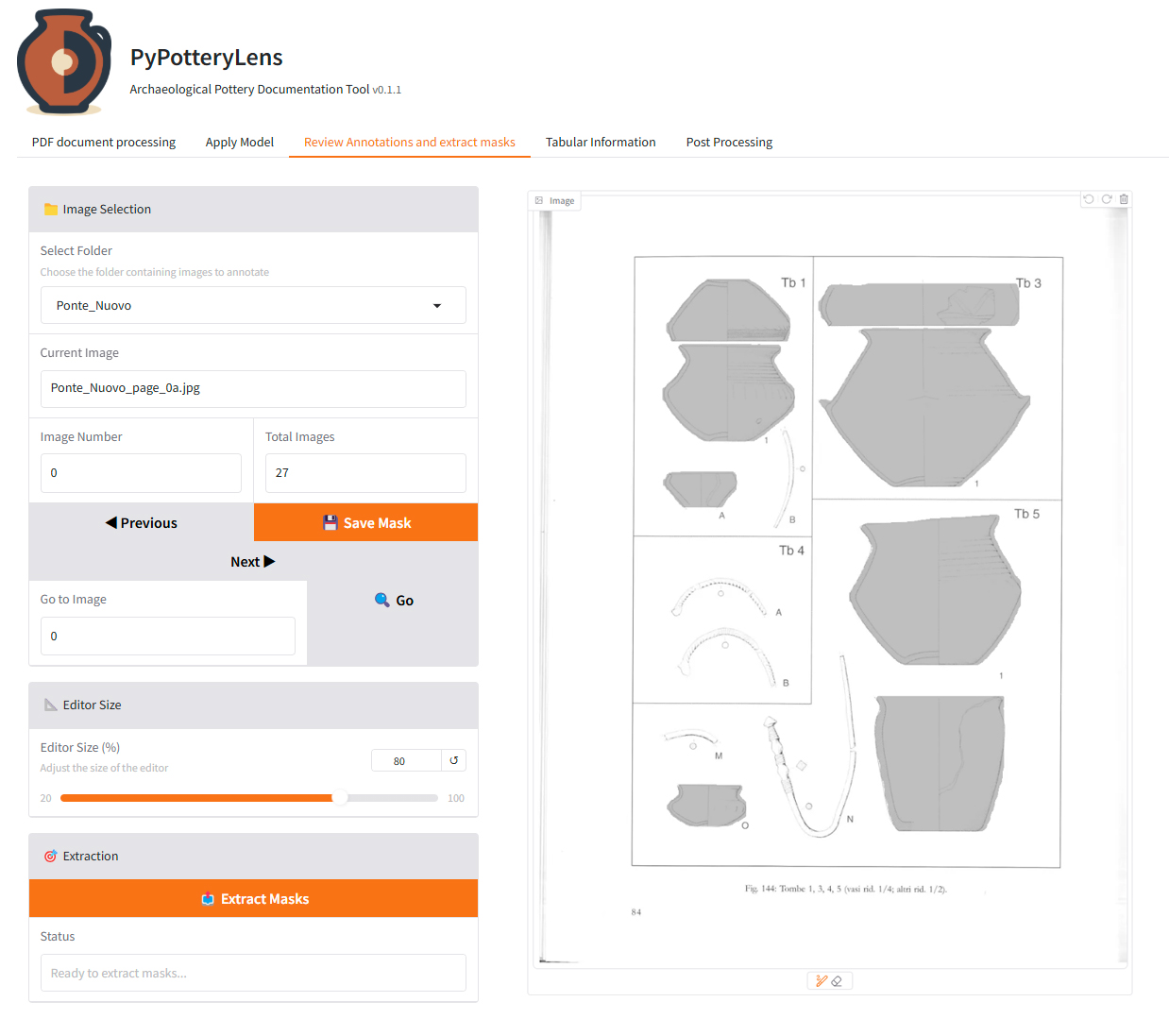}}
  
  }
  
  \caption{\label{fig-UI-example}The user interface of \emph{PyPotteryLens}. The tab for manual control of segmentation masks is shown.}
  
  \end{figure}

A key goal of \emph{PyPotteryLens} is to make advanced digital
archaeology tools accessible to all practitioners, regardless of their
technical background. While DL algorithms are traditionally developed
and deployed using programming languages like Python, I recognise that
many archaeologists may not have programming experience. This
understanding drove the decision to develop an intuitive, user-friendly
interface that bridges the gap between powerful ML capabilities and
practical archaeological workflows and knowledge (Figure~\ref{fig-UI-example}). In this sense, the
graphical user interface serves two essential purposes. First, it
democratises access to advanced analytical tools, allowing
archaeologists to focus mainly on their expertise rather than technical
implementation details. Second, it enables critical human-in-the-loop
operations, such as manual corrections and verification of results,
which are fundamental to maintaining high standards of archaeological
documentation and difficult to implement through a line of code. The UI
is built using Gradio (Abid et al. 2019), a modern web-based framework
that runs in standard web browsers. This choice ensures that
\emph{PyPotteryLens} remains platform-independent and easily accessible
across different operating systems. As you can see in the GitHub
documentation page, five of the six main modules are accessible through
the UI. The \emph{Self-Annotation} module is instead accessible via a
Python script, as it requires more specialised operations and is
intended for users with a deeper understanding of DL concepts
(Section~\ref{sec-self-annotation}).

\subsection{Harware and Software used}\label{harware-and-software-used}

The software \emph{LabelMe} (5.5.0) (Wada 2024) is used to manually
annotate the dataset. Python 3.10 is used as the main programming
language, for details on the libraries used see the
\emph{requirements.txt} file in the GitHub repository. The hardware used
for training the YOLO models is a NVIDIA L40 GPU with 24GB of VRAM
available on Google Colab. The training of the EfficientNetV2 model was
performed on a NVIDIA RTX 3070Ti GPU with 8GB of VRAM.

System requirements and compatibility are provided in the GitHub
documentation.

\section{Results}\label{results}

\subsection{Model Vision Performance}\label{sec-vision}

Since the proposed system identifies a single class, the metrics are
evaluated according to the performance in box detection and mask
segmentation. Before moving on to the analysis of the results, a brief
introduction to the metrics used should be made.

The \emph{Mean Average Precision} (\(mAP\)) measures the model's
accuracy in detecting and localising objects. It considers a detection
as correct when the \emph{Intersection over Union} (\(IoU\),
Equation~\ref{eq-iou}) between the predicted bounding box \(B\) (or
segmentation!) and the ground truth \(A\) exceeds a specific threshold.

\begin{equation}\phantomsection\label{eq-iou}{
IoU = \frac{Area_{intersection}}{Area_{union}} = \frac{|A \cap B|}{|A \cup B|}
}\end{equation}

In this analysis two key \(mAP\) variants are evaluated:

\(mAP50\) (Equation~\ref{eq-mAP50}) : This metric considers detections
to be successful if they achieve at least 50\% \(IoU\) with ground
truth.

\begin{equation}\phantomsection\label{eq-mAP50}{
mAP_{50} = \frac{1}{n}\sum_{i=1}^{n} AP_{i}^{50}
}\end{equation}

where \(AP_{i}^{50}\) is the \emph{Average Precision} at \(IoU\)
threshold of 0.5 for class \(i\)

\(mAP50-95\) (Equation~\ref{eq-mAP50-95}): This is a stricter metric
that averages \(mAP\) values across different \(IoU\) thresholds from
50\% to 95\% (in steps of 5\%).

\begin{equation}\phantomsection\label{eq-mAP50-95}{
mAP_{50:95} = \frac{1}{10}\sum_{t=0.5}^{0.95} mAP_t
}\end{equation}

where \(t\) increases in steps of 0.05

And also two classical metrics are evaluated:

\textbf{Precision} (Equation~\ref{eq-precision}): it is a measure of the
accuracy of positive predictions, calculated as the ratio of True
Positive detections to all positive predictions (True Positives + False
Positives). In other words, when the model identifies a pottery drawing,
how often is it correct?

\begin{equation}\phantomsection\label{eq-precision}{
Precision = \frac{TP}{TP + FP}
}\end{equation}

Where \(TP\) are the True Positives and \(FP\) are the False Positives.

\textbf{Recall} (Equation~\ref{eq-recall}): it measures the model's
ability to find all relevant cases, calculated as the ratio of True
Positive detections to all actual positive cases (True Positives + False
Negatives). This indicates what percentage of actual pottery drawings in
the documents are successfully detected by the model.

\begin{equation}\phantomsection\label{eq-recall}{
Recall = \frac{TP}{TP + FN}
}\end{equation}

The Table~\ref{tbl-vision-results} shows the results obtained from the
test set for the two models YOLOv8 and YOLOv11. All the metric can be
also expressed as percentage.

\begin{longtable}[]{@{}lllll@{}}
\caption{Performance metrics for YOLOv8 and YOLOv11 models on the test
set,}\label{tbl-vision-results}\tabularnewline
\toprule\noalign{}
& YOLOv8 & & YOLOv11 & \\
\midrule\noalign{}
\endfirsthead
\toprule\noalign{}
& YOLOv8 & & YOLOv11 & \\
\midrule\noalign{}
\endhead
\bottomrule\noalign{}
\endlastfoot
Metric & Box & Segmentation & Box & Segmentation \\
\(mAP50-95\) & 0.972 & 0.887 & 0.975 & 0.886 \\
\(mAP50\) & 0.99 & 0.989 & 0.989 & 0.988 \\
\(Precision\) & 0.972 & 0.972 & 0.966 & 0.963 \\
\(Recall\) & 0.967 & 0.967 & 0.967 & 0.964 \\
\end{longtable}

Both models perform very well, with all parameters above 96\% in both
box and segmentation. The only exception was the segmentation
\(mAP50-95\), which achieved approximately 89\% - still remarkably high
considering that values above 50\% are typically considered very good
performance for this strict metric (Jegham et al. 2024). At first
glance, therefore, the values obtained from the test set appear to give
similar results. For our purposes, the segmentation scores best describe
how the two models performed. Specifically, YOLOv8 has a higher
Precision than YOLOv11 (\(\approx\) 10\%), demonstrating greater
resistance to False Positives.

\begin{figure}

\centering{

\pandocbounded{\includegraphics[keepaspectratio]{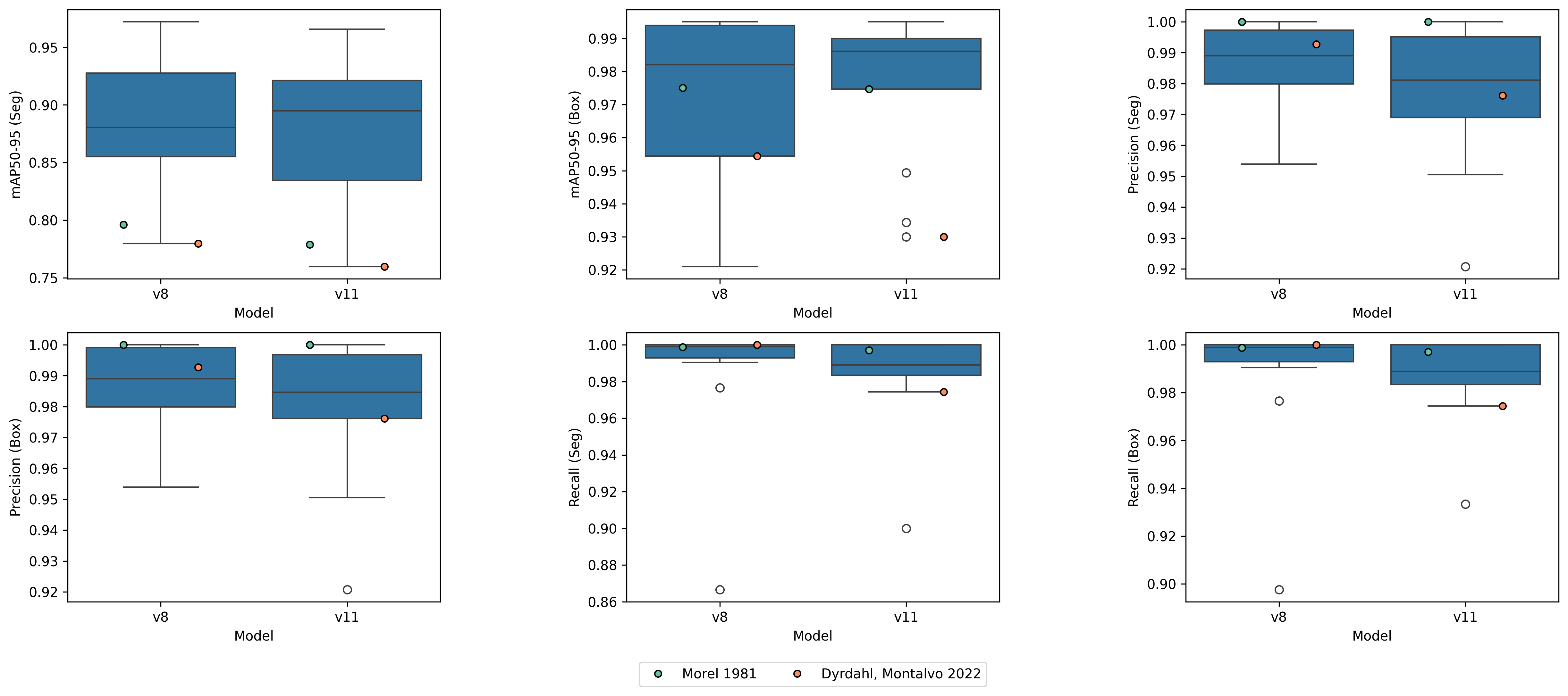}}

}

\caption{\label{fig-vision-results}Series of boxplots showing model's
validation performance metrics. Specific extra-training set contexts are
also plotted.}

\end{figure}

Finally, to ensure the model's compatibility with publications other
than those used for training, including those from different historical
periods and geographical locations, the model's behaviour is evaluated
on a test dataset composed of ceramics that differ in style and
morphology from those used for training. Tables from the Hellenistic
Roman period (Morel 1981) and a pre-Columbian site in Ecuador (Dyrdahl
and Montalvo 2022) are chosen as test cases. For all contexts, analysis
of performance distribution (Figure~\ref{fig-vision-results}) reveals
complementary strengths between the two models. While YOLOv11
demonstrates superior performance in \(mAP50-95\) metrics for both
segmentation and box detection, YOLOv8 exhibits higher Precision and
Recall values. This translates to greater resistance to False Positives
and enhanced capability in detecting ceramic instances. These
characteristics led to the selection of YOLOv8 for the final
implementation, particularly as traditional mathematical morphology
post-processing techniques (such as dilation) can effectively enhance
the quality of its segmentation masks. Moving, to the model
generalisation's capabilities, the performance metrics on the test
contexts aligned closely with training results, demonstrating robust
generalisation capabilities across different archaeological materials
(Figure~\ref{fig-vision-results}).

\subsection{Model Classifier Performance}\label{sec-classifier}

\begin{figure}

\centering{

\pandocbounded{\includegraphics[keepaspectratio]{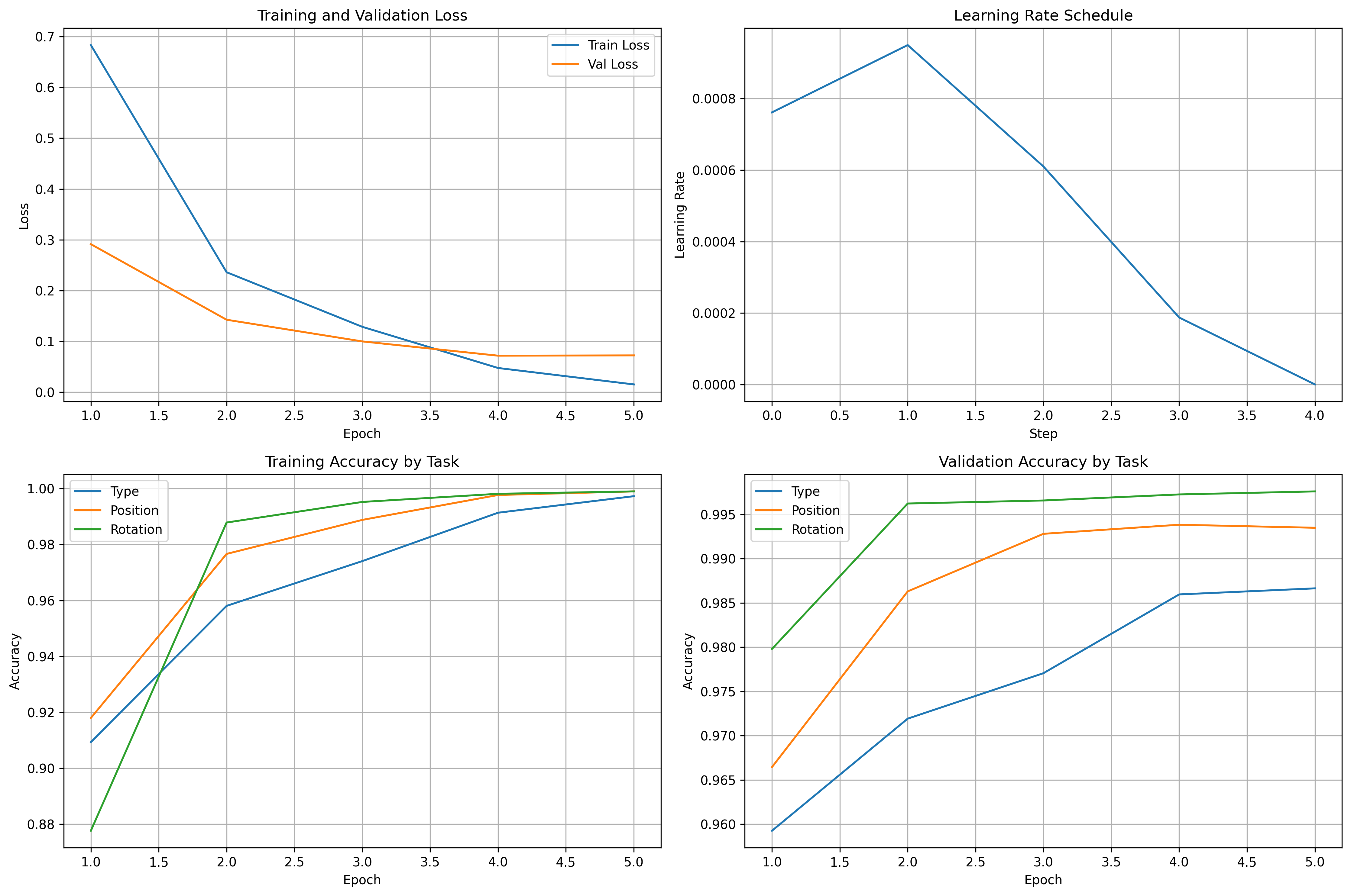}}

}

\caption{\label{fig-classifier-results}Series of plot showing different
classification performance metrics, including Loss and Accuracy.}

\end{figure}

Based on the experiments performed, the classification model based on
EfficientNetV2 performed successfully. The metrics calculated on the
validation set are offered in the following plots
(Figure~\ref{fig-classifier-results}).

Although the Figure~\ref{fig-classifier-results} shows a slight
overfitting (especially from the end of epoch 4), the accuracy on the
validation set continues to increase, reaching a value of around .995
for Rotation, .993 for Position and .986 for Type. The other metrics are
Precision and Recall, which stand at values above .98 for all three
classes (Table~\ref{tbl-classifier-results}).

\begin{longtable}[]{@{}lllllll@{}}
\caption{Performance metrics for the multi-head EfficientNetV2 model on
the validation set}\label{tbl-classifier-results}\tabularnewline
\toprule\noalign{}
& Type & & Position & & Rotation & \\
\midrule\noalign{}
\endfirsthead
\toprule\noalign{}
& Type & & Position & & Rotation & \\
\midrule\noalign{}
\endhead
\bottomrule\noalign{}
\endlastfoot
Metric & ENT & FRAG & BOTTOM & TOP & LEFT & RIGHT \\
Precision & 0.984 & 0.983 & 0.996 & 0.997 & 0.997 & 0.996 \\
Recall & 0.980 & 0.991 & 0.997 & 0.996 & 0.996 & 0.997 \\
\end{longtable}

\begin{figure}

\centering{

\pandocbounded{\includegraphics[keepaspectratio]{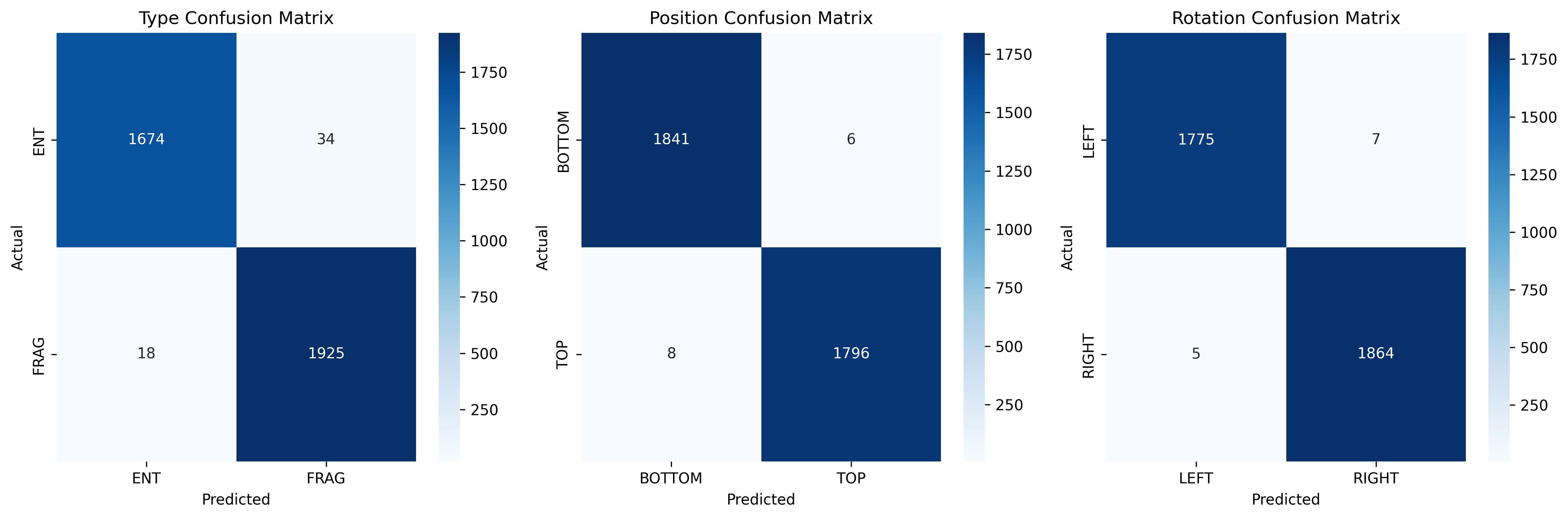}}

}

\caption{\label{fig-confusion_matrices}Confusion matrices for validation
set}

\end{figure}

In order to offer a more interpretable result, confusion matrices are
shown for each class (Figure~\ref{fig-confusion_matrices}).

Analysis of performance metrics and confusion matrices indicates that
vessel Type classification presents the highest complexity among all
classification tasks. The model exhibits a bias towards classifying
complete vessels as fragmentary, though the inverse misclassification
occurs less frequently. Despite these minor classification biases, the
overall performance metrics demonstrate robust classification
reliability across all categories.

\subsection{Case studies: Application
examples}\label{case-studies-application-examples}

In this section I provide a series of application examples of the
\emph{PyPotteryLens} system to actual archaeological contexts and
applications. The first example provides insight into the comparison
between manual and automated recording, while the second example
demonstrates the potential of the system for unsupervised analysis of
archaeological pottery and, more generally, for the integration of DL
techniques into archaeological research. All the experiments are
performed on a desktop PC with an Intel i7-12700K CPU and 32GB of RAM.

\subsubsection{Comparison with manual recording: Ponte Nuovo case
study}\label{sec-comparison}

To give a concrete idea of the time-savings compared to manual
recording, a test was carried out on an archaeological context called
Ponte Nuovo (Luciano Salzani 2005). The objective is to create drawing
instances (more traditionally, cards), associate them with bibliographic
tabular information (grave number, page, table and inventory) and create
a PDF report. In the manual process, the author tried to create points
of reference and comparison with the automatic recording, trying to
maintain a sustainable rhythm. On the other hand, the author is an
expert in manual archiving and is therefore familiar with the workflow
and keyboard shortcuts of the several programs used (GIMP, Microsoft
Excel) to speed up the process. The methods are compared on 6 main
aspects (Table~\ref{tbl-comparison}).

\begin{longtable}[]{@{}
  >{\raggedright\arraybackslash}p{(\linewidth - 4\tabcolsep) * \real{0.1463}}
  >{\raggedright\arraybackslash}p{(\linewidth - 4\tabcolsep) * \real{0.3902}}
  >{\raggedright\arraybackslash}p{(\linewidth - 4\tabcolsep) * \real{0.4634}}@{}}
\caption{Comparison of \emph{PyPotteryLens} and manual processing steps
for archaeological pottery filing}\label{tbl-comparison}\tabularnewline
\toprule\noalign{}
\begin{minipage}[b]{\linewidth}\raggedright
Step
\end{minipage} & \begin{minipage}[b]{\linewidth}\raggedright
\emph{PyPotteryLens}
\end{minipage} & \begin{minipage}[b]{\linewidth}\raggedright
Manual recording
\end{minipage} \\
\midrule\noalign{}
\endfirsthead
\toprule\noalign{}
\begin{minipage}[b]{\linewidth}\raggedright
Step
\end{minipage} & \begin{minipage}[b]{\linewidth}\raggedright
\emph{PyPotteryLens}
\end{minipage} & \begin{minipage}[b]{\linewidth}\raggedright
Manual recording
\end{minipage} \\
\midrule\noalign{}
\endhead
\bottomrule\noalign{}
\endlastfoot
1. Document Processing & PDF to image conversion & Not required (direct
PDF use) \\
2. Initial Processing & Vision model inference & Manual card creation \\
3. Cards Management & Annotation revision and mask extraction & Card
cleaning (removing non-vessel elements) \\
4. Data Entry & Tabular information input & Tabular information input \\
5. Post-processing & Automated classification using ML model & Manual
orientation correction \\
6. Final Output & Automated PDF report generation & Manual PDF report
creation \\
\end{longtable}

\begin{figure}

\centering{

\pandocbounded{\includegraphics[keepaspectratio]{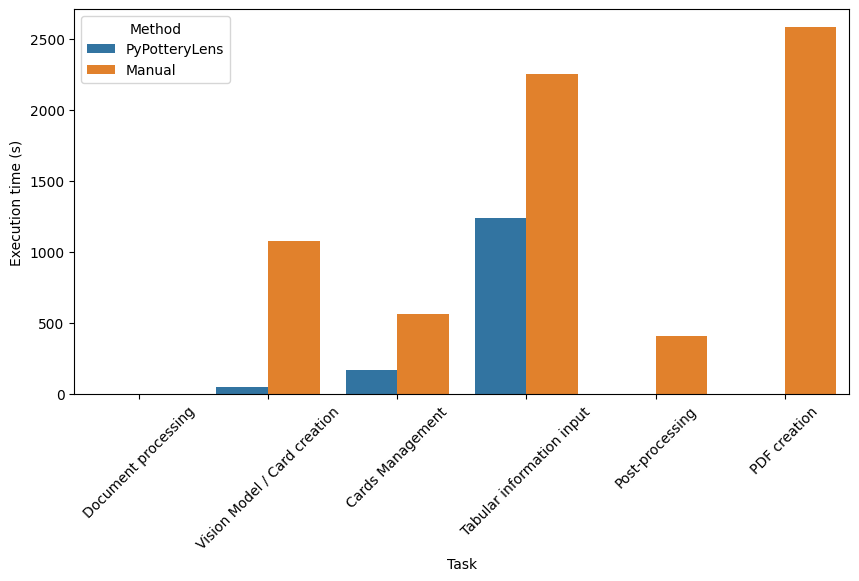}}

}

\caption{\label{fig-barplot-comparison}Barplot showing execution time
(s) for each task}

\end{figure}

The execution time (in seconds) for each step is shown in the barplot
(Figure~\ref{fig-barplot-comparison})

Excluding the Document Processing phase, which is not required for the
manual approach, the time spent on documentation is significantly higher
for the manual approach. This is somewhat predictable, but it is
interesting to note in which operation there is the greatest gain in
time. The most significant speed-ups include PDF creation: 1290× faster
(2s vs 2580s); Post-processing: 136× faster (3s vs 409s); Vision
Model/Card creation: 21× faster (51s vs 1077s). The moderate speed-ups
include Cards Management: 3.3× faster (170s vs 560s) Tabular
information: 1.8× faster (1237s vs 2252s). Simplifying,
\emph{PyPotteryLens} allowed a speed-up of 4.7× compared to manual
recording. Eliminating the Tabular information input phase results in a
speed-up of about 20×.

\newpage

\subsubsection{Deep learning integration: unsupervised analysis of Ostia
dell'Osa's pottery}\label{sec-deep-learning}

The files created through \emph{PyPotteryLens} are highly standardised,
semantically consistent and clean of other elements
(Figure~\ref{fig-seg}), making them ideal training data for DL
applications. To demonstrate this capability, I implemented an
unsupervised learning approach using a variational autoencoder (VAE)
(Kingma and Welling 2013) architecture on pottery drawings from Osteria
dell'Osa (Bietti Sestieri 1992), where over 2,300 instances are
processed through the pipeline in about half an hour. The proposed VAE
architecture combines the EfficientNetV2 (Section~\ref{sec-classifier})
as the encoder backbone with a specific designed decoder. The decoder
implements a series of transposed convolutional layers with batch
normalisation and ReLU activations, gradually reconstructing the encoded
features back into pottery images (L. Cardarelli (2022)). The loss
function used combines vanilla reconstruction loss (mean squared error)
with the Kullback-Leibler divergence term (Equation~\ref{eq-loss}):

\begin{equation}\phantomsection\label{eq-loss}{
\mathcal{L}(\theta, \phi) = \frac{1}{N}\sum_{i=1}^N (x_i - \hat{x}_i)^2 + \beta D_{KL}(q_\phi(z|x)||p(z))
}\end{equation}

where \(x_i\) represents the input image, \(\hat{x}_i\) the
reconstructed output, \(N\) is the batch size, and \(\beta\) weights the
contribution of the Kullback-Leibler divergence term \(D_{KL}\) that
regularises the latent space distribution, in this case defined as
\(0.00025\). A latent dimension of \(128\) is selected to improve
representational capacity.

\begin{figure}

\centering{

\pandocbounded{\includegraphics[keepaspectratio]{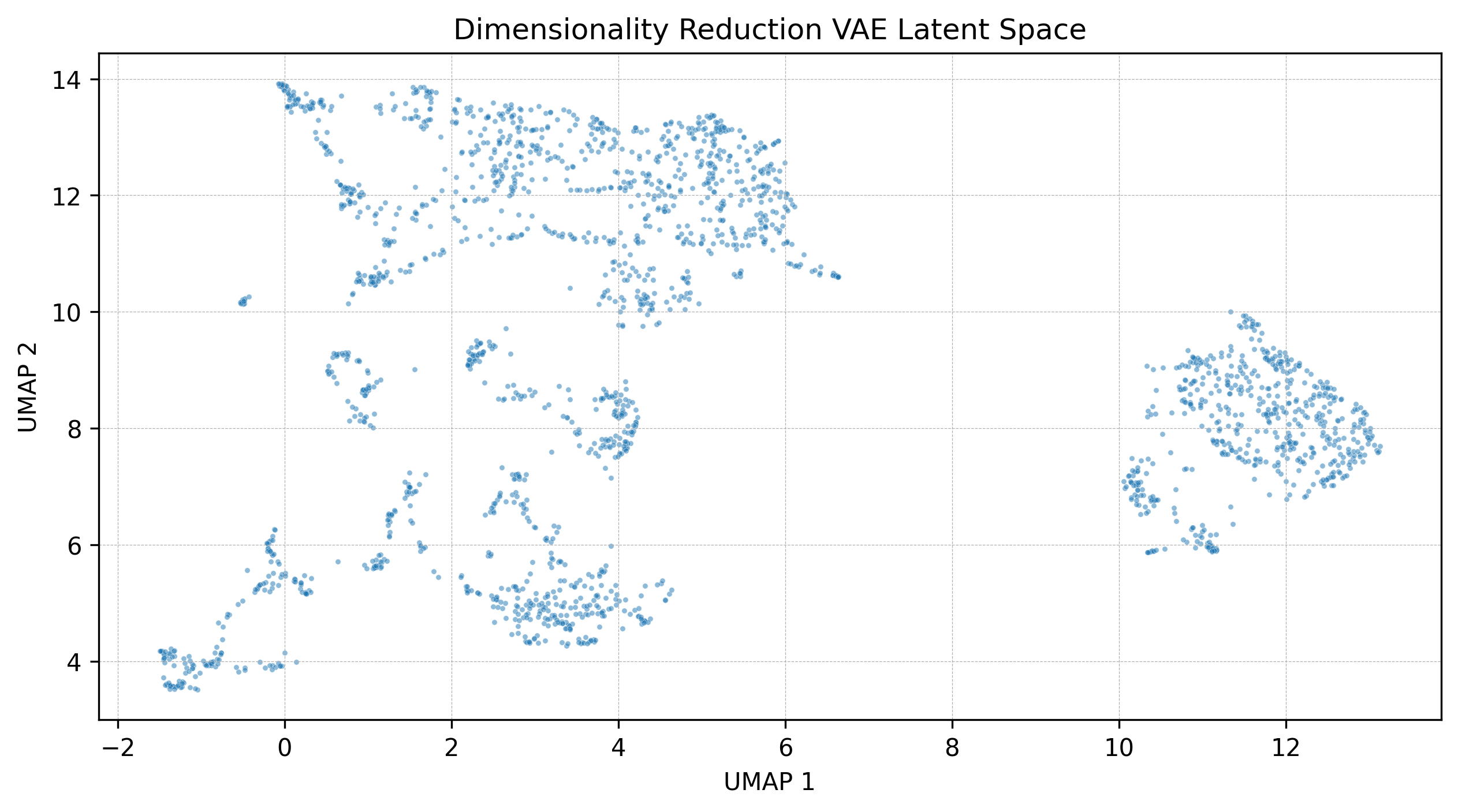}}

}

\caption{\label{fig-umap}2D embedding of the learned latent space. Clear
groupings are visible and structured in an interpretable way.}

\end{figure}

To visualise the learned latent space, I employed UMAP (Uniform Manifold
Approximation and Projection) (McInnes, Healy, and Melville 2018) in two
configurations: a 2D projection for interpretable visualisation and a 5D
embedding for more nuanced similarity analysis. The 2D projection
reveals distinct clusters or groupings corresponding to different
pottery typologies, while preserving global structure of the whole
context, providing a useful and interpretable ``map'' of the assemblage
(Figure~\ref{fig-umap}).

\begin{figure}

\centering{

\pandocbounded{\includegraphics[keepaspectratio]{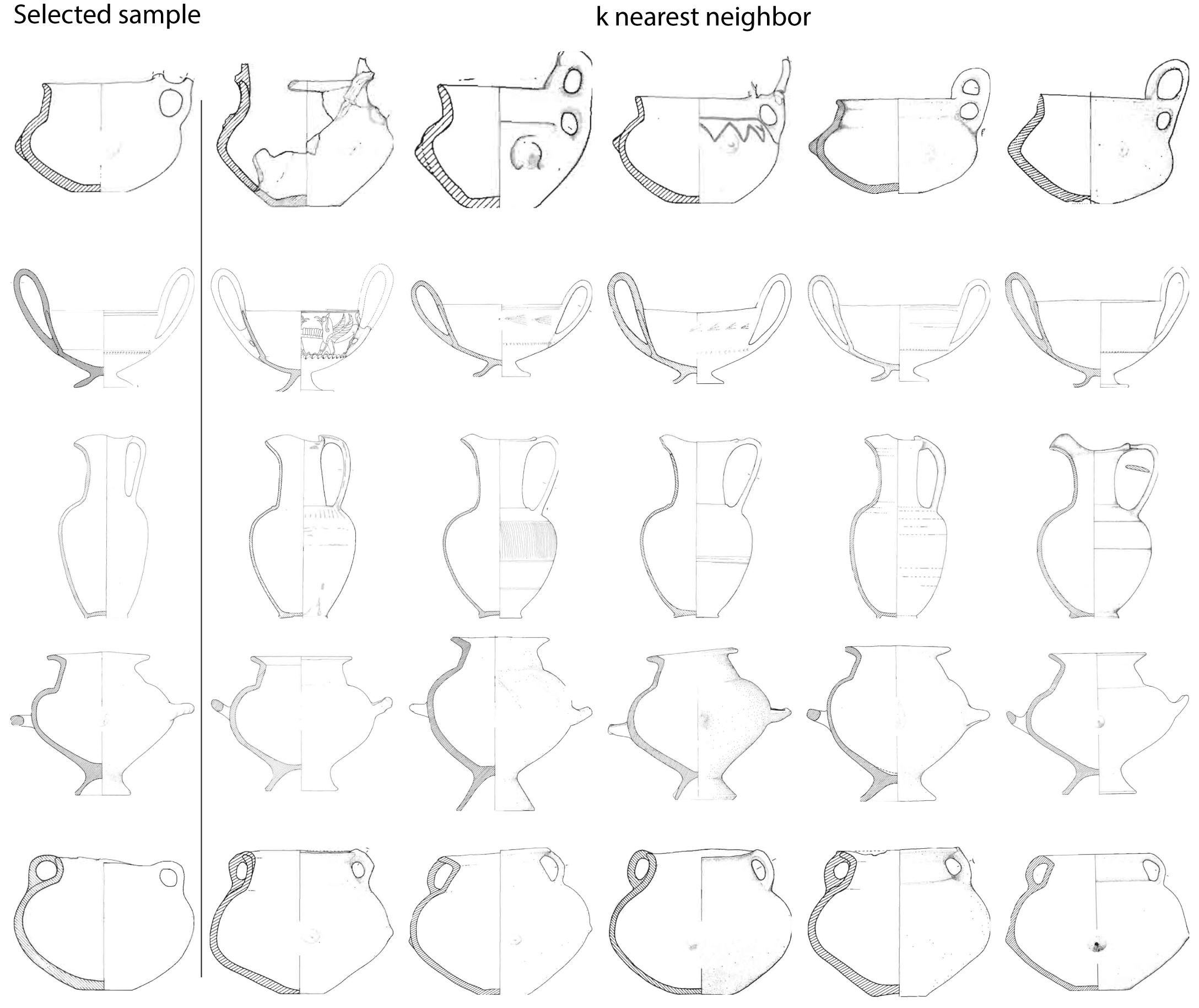}}

}

\caption{\label{fig-knn}\(k\)-nearest neighbors of selected samples.
Neighbors suggest clear morphological similarity}

\end{figure}

The higher-dimensional embedding enables more precise similarity
searches through \(k\)-nearest neighbor queries (Pinder, Shimada, and
Gregory 1979). The effectiveness of the learned representations is
validated through a qualitative analysis of nearest neighbors in latent
space. For selected query vessels, a \(k\)-nearest neighbors (\(k\)=5)
is retrived in the 5D UMAP space. The neighbors consistently exhibit
similar morphological characteristics and decorative elements,
suggesting the model learned meaningful pottery features without
explicit supervision (Figure~\ref{fig-knn}).

This capability has several significant archaeological implications,
including the rapid identification of morphological parallels across
large ceramic assemblages through unsupervised clustering and similarity
analysis, and the quantitative assessment of ceramic variation within
and between archaeological contexts. In addition, unlike previous
approaches that focus solely on vessel profiles (Navarro et al. 2021; L.
Cardarelli 2022; Parisotto et al. 2022), the method successfully
captures and encodes the full visual information present in
archaeological drawings, including decorative and handling elements.
This approach opens up a more holistic perspective to pottery analysis,
allowing the integration of multiple visual features into a single,
unified (\emph{latent}) representation.

\section{Discussion}\label{discussion}

The evaluation results (Section~\ref{sec-vision}) highlight the
capabilities of \emph{PyPotteryLens} in automating archaeological
pottery documentation and recording. The system consistently achieves
high performance metrics, with Precision and Recall values exceeding
95\% in almost all cases. Only the \(mAP50-95\) for segmentation shows
slightly lower values, though still maintaining robust and high
performance. This high accuracy extends beyond the training dataset, as
demonstrated by successful application to different archaeological
contexts, indicating strong generalisation capabilities, even in the
presence of diverse pottery styles and publication formats.

The classification model (Section~\ref{sec-classifier}) also shows high
performance, with Precision and Recall values above 98\% for all
classes. The confusion matrices show small biases in the classification
of vessel types, with a tendency to misclassify complete vessels as
fragmentary. Despite these minor biases and the low complexity of the
proposed classification task, the overall performance of the model is
robust and reliable, and the proposed multi-head architecture allows for
a good solution to multi-class classification tasks in archaeological
reality.

Regarding the time-saving objective (Section~\ref{sec-comparison}), The
Ponte Nuovo case study effectively illustrates the practical advantages
of this automated approach. The significant reduction in processing time
compared to manual documentation represents a substantial improvement in
workflow efficiency. However, it's important to note that automation
does not entirely eliminate the need for archaeological expertise.
Rather, it transforms the role of the archaeologist from manual
processor to expert validator, particularly in tasks requiring
specialist knowledge and interpretation, such as chronological or
typological classification. In fact, from an archaeological point of
view, the impact of the system goes beyond mere time savings (although
it does offer the possibility of increasing the sample size, for
example). The automation of initial documentation and classification
tasks allows archaeologists to focus more on interpretation and
analysis, enhancing the quality of archaeological research. The
inclusion of manual correction tools for both vision and classification
models ensure that archaeological expertise remains a pivot point in the
documentation process. A key challenge emerges in the processing of
tabular information, where time savings are less pronounced compared to
other tasks (Figure~\ref{fig-barplot-comparison}). This limitation stems
from the inherent structure of archaeological publications, where
contextual information is often presented in unstructured formats and
may be physically separated from the associated images (e.g., in another
chapter). While Large Language Models (LLMs) could potentially address
this challenge, their implementation would require substantial
computational resources. Another potential solution could involve the
development custom YOLO models to detect page layouts and extract useful
information (Zhao et al. 2024), speeding-up the tabular information
input phase especially regarding bibliographic references input.

The Osteria dell'Osa example (Section~\ref{sec-deep-learning}),
demonstrates the system's potential as a base for further computational
analysis. In this case, the data collection phase is completely
automated. By analysing full images, our training data is generated from
the system itself, eliminating a long and tedious documentation process.
While a demonstration, the results are promising: using large training
dataset the analysis can be expanded allowing for more complex and
detailed analysis that include also decoration and handling elements. As
a drawback, the use of images can be a limitation in the case of very
different publication styles (Figure~\ref{fig-pottery_style}), where the
system may struggle to generalise and focus on the style of the
publication rather than the pottery itself. In this case, the use of a
more complex model, such as a vision transformer (Ramachandran et al.
(2019); Dosovitskiy et al. (2021)) or style-transfer (Dumoulin, Shlens,
and Kudlur (2017); Jing et al. (2018); Zhu et al. (2020)), could be a
solution, but it would require a more complex training process and a
more powerful hardware. After all, more than 2,000 images were analysed
in just over half an hour, making this approach a powerful exploratory
tool for archaeological analysis, allowing subsequent analyses to be
more targeted and detailed. In this scenario, Data is power. What we can
do with it is limited only by our imagination and our research
questions.

Moving back to general considerations, the current focus on pottery,
while potentially viewed as a limitation, actually demonstrates the
system's potential for expansion. \emph{PyPotteryLens}'s modular
architecture allows for adaptation to other material classes through
targeted model training and vision model selection. It's enough to train
a new model on a dataset of a different material class, and the system
can be easily adapted to a new task. The ability to process a single
class of material can be seen as a weakness, as it limits the system's
versatility. However, this choice was made to ensure the system's
robustness and reliability in single-class documentation tasks, ensuring
a straightforward and easy to develop solution.

Publishing the source code and documentation on GitHub ensures that the
system is accessible and adaptable to the archaeological community,
facilitating the development of new models, features and the application
of the system to other materials.

\section{Conclusion and Future Work}\label{conclusion-and-future-work}

This study introduces \emph{PyPotteryLens}, demonstrating how
specialised DL approaches can transform archaeological pottery
documentation while preserving its essential analytical value. The
framework achieves exceptional performance metrics, reducing processing
time by up to 20× compared to manual methods while maintaining high
Precision and Recall rates across diverse archaeological contexts.
Through the analysis of over 2,300 pottery instances from the Osteria
dell'Osa assemblage, the system demonstrates remarkable capabilities for
rapid and sophisticated unsupervised learning analysis, opening new
pathways for exploring typological relationships and performs
morphological analysis in standardised digital archives. The seamless
integration of computer vision techniques with archaeological expertise
creates a powerful tool that not only accelerates documentation
processes but also enriches analytical capabilities through standardised
data generation. Significantly, the framework's modular, open-source
architecture establishes a foundational methodology for computational
archaeology that prioritises reproducibility, community engagement, and
accessibility. Looking toward future developments, three key areas of
improvement emerge as particularly promising: (1) Enhanced
generalization through style-transfer approaches to accommodate the rich
diversity of publication formats and drawing styles, particularly for
autoencoding tasks; (2) Automated metadata extraction utilizing natural
language processing and specialized YOLO-based models to capture
contextual information from publication text; and (3) Expanded material
coverage to adapt the framework for other archaeological materials,
including lithics and metallurgical artifacts, while maintaining
methodological consistency.

\section{Data Availability}\label{data-availability}

The software, documentation, and examples are available on \href{https://github.com/lrncrd/PyPottery/tree/PyPotteryLens}{GitHub}, while models are availabe on \href{https://huggingface.co/lrncrd/PyPotteryLens/tree/main}{HuggingFace}. Due to
copyright restrictions, the training dataset is not publicly available.
Collaborations for the development of new datasets and software's
features are welcome.

\section{Acknowledgements}\label{acknowledgements}

The author would like to thank the annotators who contributed to the
training dataset: Annalisa Lapadula, Dacia Cardarelli, Lucrezia
Petrucci. The author would also like to thank Bruno Papa for the support
in the development of the software.

\newpage{}

\section*{References}\label{references}
\addcontentsline{toc}{section}{References}

\phantomsection\label{refs}

\end{document}